%% file: main.tex
\PassOptionsToPackage{numbers,sort&compress}{natbib}
\PassOptionsToPackage{hyphens}{url}
\PassOptionsToPackage{shortlabels,inline}{enumitem}
\documentclass[10pt,twocolumn,letterpaper]{article}

\input{preamble}

\usepackage[algorithms]{wacv}  

\definecolor{wacvblue}{rgb}{0.21,0.49,0.74}

\def\wacvPaperID{2720} 
\def\confName{WACV}
\def\confYear{2026}

\title{Uncertainty-Aware Subset Selection for Robust Visual Explainability under Distribution Shifts}

\author{
Madhav Gupta \quad
Vishak Prasad \quad
Ganesh Ramakrishnan\\[4pt]
\textnormal{Indian Institute of Technology Bombay, India}\\
{\tt\small 21d070043@iitb.ac.in \quad vishak@cse.iitb.ac.in \quad ganesh@cse.iitb.ac.in}
}

\begin{document}

\maketitle

\begingroup
\renewcommand\thefootnote{}\footnotetext{Link to the code - \href{https://github.com/madhavgupta6803/Visual-Explainability-under-Distribution-Shifts}{Visual Explainability under Distributional Shifts}}
\addtocounter{footnote}{-1}%
\endgroup

\begin{abstract}

Subset selection–based methods are widely used to explain deep vision models: they attribute predictions by highlighting the most influential image regions and support object-level explanations. While these methods perform well in in-distribution (ID) settings, their behavior under out-of-distribution (OOD) conditions remains poorly understood. Through extensive experiments across multiple ID–OOD sets, we find that reliability of the existing subset based methods degrades markedly, yielding redundant, unstable, and uncertainty-sensitive explanations. To address these shortcomings, we introduce a framework that combines submodular subset selection with layer-wise, gradient-based uncertainty estimation to improve robustness and fidelity without requiring additional training or auxiliary models. Our approach estimates uncertainty via adaptive weight perturbations and uses these estimates to guide submodular optimization, ensuring diverse and informative subset selection. Empirical evaluations show that, beyond mitigating the weaknesses of existing methods under OOD scenarios, our framework also yields improvements in ID settings. These findings highlight limitations of current subset-based approaches and demonstrate how uncertainty-driven optimization can enhance attribution and object-level interpretability, paving the way for more transparent and trustworthy AI in real-world vision applications.

\end{abstract}

\section{Introduction}


Deep vision models are increasingly deployed in safety-critical and user-facing applications, such as autonomous driving and medical imaging, where interpretability is essential for building user trust, facilitating debugging, and complying with regulatory and safety requirements \cite{explain_autonomous_driving, steex_eccv2022, framework_medical_imaging, xai_regulation}. Visual attribution which is the practice of highlighting input regions most responsible for a model’s prediction, has thus become a core research area, spanning gradient-based methods such as Grad-CAM and Integrated Gradients \cite{selvaraju2017grad,sundararajan2017axiomatic}, and perturbation-based approaches like occlusion, Meaningful Perturbations, and RISE \cite{zeiler2014visualizing, fong2017meaningful, petsiuk2018rise}.
Despite these advances, existing attribution methods suffer from notable pitfalls: saliency maps can be insensitive to model parameters or inputs, and explanations can be brittle and change sharply under small image or weight perturbations \cite{zhang2025empirical, etmann2019robustness} undermining their reliability for auditing and debugging \cite{adebayo2018sanity}. The challenge becomes even more pronounced under distribution shift \cite{farquhar2022ood}: confidence estimates often degrade and uncertainty becomes unreliable, making post-hoc explanations on OOD inputs unreliable without robustness safeguards \cite{ovadia2019can}. For example, a model predicting the label “Cat” should consistently highlight distinctive features such as ears and whiskers, even when the input deviates from the training distribution; for example a dog, yet existing attribution methods fail to do so under such shifts.
Attribution now underpins multiple application domains, including model transparency and interpretability frameworks like LIME \cite{ribeiro2016lime}, SHAP \cite{lundberg2017shap} and more recent dependence-based approaches such as HSIC Attribution \cite{novello2022hsic}, which leverages the Hilbert-Schmidt Independence Criterion to quantify the statistical dependence between input features and model outputs. These methods are increasingly applied in vision-language systems where identifying key patches aids alignment, medical imaging for anomaly localization, and robustness analysis.

A promising line of research frames explanation as subset selection over image regions to identify the most informative evidence while maintaining interpretability; examples include maximizing submodular function for subset selection \cite{chen2024less}, deletion/insertion masks on subsets for metric driven attribution \cite{walker2023metric}, and visual precision search \cite{chen2025visualprecision}. However, current subset-selection approaches often over-select redundant patches, and are optimized for in-distribution data, leading to poor generalization under distribution shifts.
Motivated by these gaps, our work explores how to make attributions reliable under distribution shift by integrating submodular subset selection with principled uncertainty estimation via adaptive weight perturbations, drawing inspiration from Bayesian \cite{warburg2023bayesian} and ensemble approximations such as Deep Ensembles \cite{lakshminarayanan2017deep}, and SWAG \cite{maddox2019swag}.

Our approach combines submodular subset selection with adaptive uncertainty estimation derived from weight perturbations, leveraging gradient sensitivity, adaptive noise scaling, and Mahalanobis-based normalization to prioritize informative and stable image regions \cite{huang2021importance, lee2018simple}. This integration is lightweight, requiring no additional training or uncertainty models, and is easily adaptable across architectures.

To rigorously evaluate attribution robustness under distribution shift, we curated paired in-distribution (ID) and out-of-distribution (OOD) datasets following principles outlined in \cite{farquhar2022ood}. For each ID dataset, the OOD counterpart was designed to be related, complementary, and transformed, ensuring the shift was realistic and semantically meaningful rather than arbitrary. Across these carefully curated ID-OOD pairs, we observed that existing subset selection attribution methods perform well in ID settings but degrade severely in OOD scenarios, with insertion and deletion scores dropping up to 40\%. In contrast, our uncertainty-aware framework significantly closes this robustness gap while also improving ID fidelity, demonstrating that uncertainty-driven subset selection is crucial for interpretable attribution under distributional shift.
Overall our key contributions are:
\begin{itemize}
    \item We empirically show that existing subset-selection methods perform poorly under distribution shift, revealing a critical robustness gap.
    \item We propose a novel attribution framework that integrates submodular optimization with adaptive uncertainty estimation to prioritize stable and informative regions.
    \item Our method operates using only a fine-tuned backbone, making it lightweight and generalizable across architectures and datasets.
    \item Our approach improves the evaluation metric scores in ID scenarios while closing the robustness gap on OOD datasets.
\end{itemize}

\section{Related Works}
\label{sec:related}

Our work sits at the intersection of four research threads: gradient-based and propagation-based visual attribution, subset selection based methods for interpretable explanations, uncertainty and weight-space methods for OOD detection, and robustness benchmarks for transformed or corrupted inputs.

\textbf{Gradient-based attribution.} Classic methods such as Grad-CAM \cite{selvaraju2017grad}, Integrated Gradients \cite{sundararajan2017axiomatic}, and LRP \cite{bach2015lrp} highlight input features via gradients or relevance propagation. While simple and general, these maps are often noisy and brittle under shifts \cite{adebayo2018sanity}. Rather than producing dense saliency, we derive layerwise gradient norms under adaptive perturbations to obtain an uncertainty-aware confidence signal that guides selection.  

\textbf{Subset selection.} Region-based attribution via patch/subset selection yields compact and interpretable explanations \cite{chen2024less, chen2025visualprecision, walker2023metric}. Submodular formulations in particular encourage diversity and efficiency, but they degrade under OOD. We address this by augmenting the objective with gradient-based uncertainty, stabilizing greedy search under distribution shift.  

\textbf{Uncertainty and ensembles.} Deep Ensembles \cite{lakshminarayanan2017deep} and SWAG \cite{maddox2019swag} exemplify strong epistemic uncertainty methods, while SGD-noise variants (e.g., S-SGD \cite{sung2020ssgd}) explore flat minima. Inspired by these, we adopt lightweight weight perturbations at inference, forming a Monte Carlo ensemble without retraining.  

\textbf{Feature-space OOD detectors.} Mahalanobis-style detectors fit Gaussian class-conditional features and measure confidence via distance \cite{fort2021limits, lee2018simple}. We extend this idea to \emph{gradient descriptors}, giving a normalized sensitivity score that captures both subtle and large OOD deviations.  

\textbf{Corruptions and Transformed Inputs.} Benchmarks such as ImageNet-C highlight performance drops under noise, blur, and geometric transformations \cite{hendrycks2019benchmarking}. Our adaptive scaling of perturbations based on layerwise feature deviation explicitly targets this regime and, together with layerwise aggregation, differentiates early-layer corruption effects from deeper semantic changes.

\textbf{How we differ.} Our method differs in three key ways: (i) we compute \emph{layerwise gradient norms under input- and layer-aware weight perturbations}, producing descriptors that encode sensitivity rather than raw activations; (ii) our perturbations are \emph{adaptive} to both the layer statistics and the input’s deviation from a penultimate centroid, which helps separate complementary, related, and transformed OOD and (iii) we integrate the resulting uncertainty scores directly into a submodular selection objective, thereby producing attribution subsets that are robust to OOD while remaining compact and interpretable. We therefore bridge uncertainty-aware OOD detection and subset-selection based attribution in a lightweight, plug-and-play manner.

\bigskip

We note that several recent works combine uncertainty or ensemble signals with attribution or selection mechanisms; where parallel ideas exist we have highlighted the distinctions above and emphasized the contributions of adaptive, gradient-derived descriptors and their integration into submodular optimization.

\section{Methodology}
\subsection{Background}
\begin{figure}[ht]
    \centering
    \begin{minipage}[b]{0.4\linewidth}
        \includegraphics[height=3.5cm]{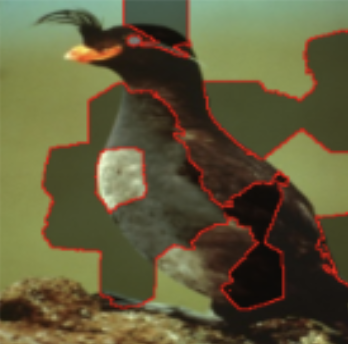}
        \caption{\textbf{Attribution map for an ID sample}: Semantically coherent regions highlight the bird's key features}
        \label{fig:CUB_Sample}
    \end{minipage}
    \hspace{0.01\linewidth}
    \begin{minipage}[b]{0.55\linewidth}
        \includegraphics[height=3.5cm]{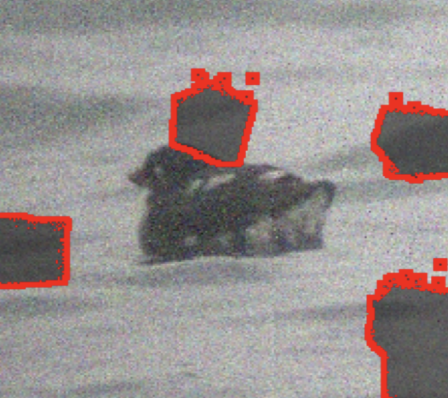}
        \caption{\textbf{Attribution map for an OOD sample}: Quality of selected subsets degrades as they become fragmented and show irrelevant background}
        \label{fig:NA_Birds_Sample}
    \end{minipage}
\end{figure}

Visual attribution methods aim to highlight the most informative image regions contributing to a model’s prediction. Subset selection-based attribution \cite{chen2024less, chen2025visualprecision} has shown promise on In-Distribution (ID) data, such as bird species from the CUB dataset, where the selected subsets emphasize semantically meaningful features like the eye, beak shape, and head patterns (Figure:~\ref{fig:CUB_Sample}). These features are crucial for fine-grained species classification and result in clear, interpretable explanations.


\begin{figure*}[ht]
\centering
\includegraphics[width=1\linewidth]{Screenshot_2025-09-20_at_1.36.13_PM.png}
\caption{\textbf{An overview of the uncertainty-aware submodular selection framework}. The process evaluates candidate image patches, generated by a baseline method, to select the most informative subset for an explanation. This selection is guided by a submodular objective function where all 4 component scores-Confidence, Effectiveness, Consistency and Collaboration, are derived from a single \textbf{Recognition Model}. Our novel \textbf{Confidence Score} is calculated by measuring the model's output stability under adaptive noise perturbations. The framework greedily selects a compact and reliable explanation, with its final quality assessed by metrics like the Insertion AUC score (right).}
\label{fig:Architecture}
\end{figure*}

However, when the same method is applied to Out-of-Distribution (OOD) samples, such as species from the North American Birds dataset which is a related OOD to CUB dataset \cite{farquhar2022ood}(Figure:~\ref{fig:NA_Birds_Sample}), the attribution quality degrades significantly. The selected subsets often include irrelevant background regions or fragmented patches, failing to capture distinctive characteristics of the bird. This lack of consistency under distribution shift indicates that current subset selection approaches overfit to ID characteristics and struggle to generalize to unseen domains.

The root causes include reliance on confidence-based scoring, lack of gradient-informed based uncertainty modeling, and absence of diversity constraints, making these methods brittle under OOD scenarios.

The goal of this work is to develop a robust attribution framework that generalizes across ID and OOD data without requiring additional trained uncertainty estimation models. To achieve this, we propose leveraging uncertainty-aware submodular optimization to identify stable and informative subsets even under distributional shifts. Our approach ensures attribution fidelity and robustness by consistently highlighting critical semantic features across both ID and OOD scenarios, thereby reducing redundancy and improving interpretability. Recognizing that different interpretation tasks require tailored objectives, we propose two distinct uncertainty-aware submodular functions: the first for robust visual attribution in general classification tasks, and the second for robust object-level interpretation of modern foundation models. Both functions are designed to improve attribution fidelity and robustness under distribution shifts without requiring specialized model training. The core of our contribution is a versatile confidence score that can be integrated into different submodular compositions to make them resilient to OOD conditions.

\subsection{Uncertainty-Aware Submodular Objective Functions}

We frame visual interpretation as a problem of maximizing an objective function $F(S)$ over a subset of image regions $S$. We propose two specific formulations of $F(S)$ tailored to different tasks.

\subsubsection{Objective Function for Robust Visual Attribution}

For the task of general visual attribution, we build upon the framework proposed by Chen et al. in \cite{chen2024less}. We adapt their objective function structure, which combines scores for effectiveness, consistency, and collaboration. However, we replace their original confidence term with our novel gradient-based uncertainty score derived confidence to explicitly enhance OOD robustness. The resulting objective function, $F_{\text{attr}}(S)$, is:
\begin{equation}
\begin{split}
F_{\text{attr}}(S) = {} & \mu_1 s_{\text{conf}}(S) + \mu_2 s_{\text{eff}}(S) \\
& + \mu_3 s_{\text{cons}}(S, f_s) + \mu_4 s_{\text{colla}}(S, I, f_s)
\end{split}
\label{eq:objective_function_attr}
\end{equation}

Here, $s_{\text{eff}}$, $s_{\text{cons}}$, and $s_{\text{colla}}$ are adopted from Chen et al. \cite{chen2024less} . Effectiveness Score promotes diversity by discouraging the selection of semantically redundant regions. The Consistency Score ensures relevance by selecting regions whose features are semantically aligned with the target class being explained and the Collaboration Score measures the synergistic importance of regions. We replace the original confidence term in this framework with our novel gradient-based uncertainty score to explicitly enhance robustness to distributional shifts. The key component, $s_{\text{conf}}$, is our proposed uncertainty-aware score, detailed in Section~\ref{sec:confidence_score}.

\subsubsection{Objective Function for Robust Object-Level Model Interpretation}

For the specific challenge of interpreting object-level foundation models (e.g., for visual grounding or object detection), we propose a second objective function, $F_{\text{obj}}(S)$. This function integrates our uncertainty measure into the score composition from the Visual Precision Search (VPS) method \cite{chen2025visualprecision}. The original VPS framework introduces a Clue Score to identify regions supporting accurate object localization and a Collaboration Score is used to assess the collective importance of a subset by measuring the drop in detection performance when those regions are removed from the image. We incorporate both along with our proposed gradient aware confidence score $s_{\text{conf}}$ to make the search process more robust.

By augmenting these two scores with our uncertainty-aware confidence term, we construct our second objective function:
\begin{equation}
\begin{split}
F_{\text{obj}}(S) = {} & \mu_1 s_{\text{conf}}(S) 
+ \mu_2 s_{\text{clue}}(S, b_{\text{target}}, c) \\
& + \mu_3 s_{\text{colla-obj}}(S, b_{\text{target}}, c)
\end{split}
\label{eq:objective_function_obj}
\end{equation}

This formulation directs the search to find regions that are not only important for the detection task (high clue and collaboration) but also yield a confident and stable prediction from the model (high confidence), making the resulting interpretation more trustworthy.

\subsection{Core Component: Gradient-Based Confidence Score ($s_{\text{conf}}$)}
\label{sec:confidence_score}

Our primary contribution, the confidence score $s_{\text{conf}}$, is a versatile module used in both $F_{\text{attr}}(S)$ and $F_{\text{obj}}(S)$. It is designed to quantify model uncertainty by assessing the stability of a model's output with respect to stochastic perturbations of its parameters. This approach is grounded in the principle of simulating epistemic uncertainty, which arises from limited knowledge of model parameters, by creating a localized ensemble of models in weight space and measuring the resulting variance in the model’s input–output sensitivity. The full procedure is detailed in Algorithm~\ref{alg:uncertainty}.

The use of gradient norms as a confidence proxy \cite{lee2020gradients} is based on the intuition that when a model is uncertain, its output is more sensitive to small changes in the input. Thus, larger gradient norms \cite{zhao2022gradnorm} signal higher input sensitivity \cite{novak2018sensitivity} and, correspondingly, lower prediction confidence \cite{riedlinger2021gradient}. This makes the gradient norm a meaningful and informative signal for quantifying uncertainty. Moreover, the dynamic scaling of noise \cite{li2024noise} based on layer-specific weight magnitudes ensures that the perturbations are proportionally significant across different layers of the network. This guards against under- or over-perturbation in layers with vastly different parameter distributions, enhancing the reliability of the uncertainty estimation.

Unlike traditional fixed-noise approaches \cite{sung2020ssgd}, our method introduces adaptivity into the noise injection process, ensuring that perturbations are relevant to each layer’s scale and functional role. This enables more realistic simulations of model variability and enhances the interpretability of resulting confidence scores.

Furthermore, the proposed approach operates independently of softmax-based confidence measures \cite{pearce2021understanding} and does not rely on specialized training procedures such as those involving Dirichlet distributions or auxiliary calibration models. This makes it broadly applicable across tasks and architectures, including in out-of-distribution (OOD) scenarios \cite{xie2024gdscore} where traditional confidence measures tend to perform poorly.

\begin{algorithm}[t]
\caption{Layer-wise Gradient \& Mahalanobis Uncertainty Estimation}
\label{alg:uncertainty}
\SetAlgoLined
\KwIn{Input batch $\mathbf{X} = \{\mathbf{x}_i\}_{i=1}^B$, model $\mathcal{M}$ with layers $\{\ell_1, \dots, \ell_K\}$, 
training penultimate features $\Phi_{\text{train}}$, training descriptors $D_{\text{train}}$, 
number of stochastic passes $T$, base noise scale $\alpha$, adaptive parameters $\beta, \gamma$, ridge $\lambda$.}
\KwOut{Normalized uncertainty scores $\mathbf{u} = \{u_i\}_{i=1}^B$}

Compute penultimate centroid: $\bar{\phi} \gets \text{mean}(\Phi_{\text{train}})$\;
Compute median distance: $\rho_0 \gets \text{median}(\|\Phi_{\text{train}} - \bar{\phi}\|_2)$\;

Compute Mahalanobis parameters if descriptors available: 
$\mu \gets \text{mean}(D_{\text{train}})$, 
$\Sigma \gets \text{cov}(D_{\text{train}}) + \lambda I$\;

\For{$t = 1, 2, \dots, T$}{
    \For{each layer $\ell \in \{\ell_1, \dots, \ell_K\}$}{
        Compute weight statistics: $\sigma_\ell \gets \text{std}(\theta_\ell)$\;
        Compute adaptive scaling $u(\mathbf{x}) \gets 1 + \beta \cdot \tanh(\gamma (\|\phi_\ell(\mathbf{x}) - \bar{\phi}\|_2 - \rho_0))$\;
        Sample Gaussian noise: $\epsilon_\ell \sim \mathcal{N}(0,1)$\;
        Perturb weights: $\tilde{\theta}_\ell^{(t)} \gets \theta_\ell + \alpha \cdot \sigma_\ell \cdot u(\mathbf{x}) \cdot \epsilon_\ell$\;
    }
    Forward pass: $(\mathbf{a}_1^{(t)}, \dots, \mathbf{a}_K^{(t)}, \hat{y}^{(t)}) \gets \mathcal{M}(\mathbf{X}; \tilde{\theta}^{(t)})$\;
    \For{each layer $\ell_k$}{
        Compute gradient: $\mathbf{g}_{\ell_k}^{(t)} \gets \nabla_{\mathbf{a}_k} \hat{y}^{(t)}$\;
        Compute layer-wise norm: $\eta_{\ell_k}^{(t)} \gets \|\mathbf{g}_{\ell_k}^{(t)}\|_2$\;
    }
}
Aggregate descriptors across $T$ passes: $\mathbf{d}_i \gets \frac{1}{T} \sum_{t=1}^T [\eta_{\ell_1}^{(t)}, \dots, \eta_{\ell_K}^{(t)}]$\;

Compute Mahalanobis / distance score for each sample:
\[
s_i \gets 
\begin{cases} 
\sqrt{ (\mathbf{d}_i - \mu)^T \Sigma^{-1} (\mathbf{d}_i - \mu) }, & \text{if } D_{\text{train}} \text{ available} \\
\|\mathbf{d}_i - \text{mean}(\mathbf{d})\|_2, & \text{otherwise}
\end{cases}
\]

Normalize scores to $[0,1]$: $u_i \gets (s_i - \min_j s_j)/(\max_j s_j - \min_j s_j)$\;

\Return $\mathbf{u}$\;
\end{algorithm}

\subsection{Algorithm Description }

We propose a gradient-based uncertainty estimator that injects input-aware, layer-wise stochastic perturbations into network weights. For a batch $\mathbf{X}=\{\mathbf{x}_i\}_{i=1}^B$, each layer weight $\theta_\ell$ is perturbed with Gaussian noise $\epsilon_\ell\!\sim\!\mathcal{N}(0,1)$ as
\begin{equation}
\tilde{\theta}_\ell^{(t)} \;=\; \theta_\ell + \alpha\,\sigma_\ell\,u(\mathbf{x})\,\epsilon_\ell,
\label{eq:weight_perturb}
\end{equation}
where $\sigma_\ell=\mathrm{std}(\theta_\ell)$ captures layer-scale statistics and $\alpha$ is a global scale. The input-aware modulation is defined by
\begin{equation}
u(\mathbf{x}) \;=\; 1 + \beta\;\tanh\!\big(\gamma(\|\phi_\ell(\mathbf{x})-\bar{\phi}\|_2 - \rho_0)\big),
\label{eq:input_mod}
\end{equation}
with $\phi_\ell(\mathbf{x})$ denoting layer features, $\bar{\phi}$ the training centroid, and $\beta,\gamma,\rho_0$ controlling amplitude, sensitivity and thresholding of the modulation respectively.

The adaptive design is motivated by the heterogeneous roles of network layers i.e. early layers predominantly encode low-level statistics (e.g., textures, edges) while deeper layers capture higher-level semantics \cite{zeiler2014visualizing, yosinski2015understanding, raghu2017svcca}. Applying a uniform perturbation across layers therefore risks two failure modes: (i) \emph{under-perturbing} deep semantic layers so that semantically significant deviations remain undetected, and (ii) \emph{over-perturbing} sensitive shallow layers which amplifies irrelevant, high-frequency noise and yields unstable attribution signals. The adaptive term $\sigma_\ell u(\mathbf{x})$ counters these issues by scaling noise proportionally to layer statistics and to the input’s distance from the training manifold: when $\mathbf{x}$ is near $\bar{\phi}$, $u(\mathbf{x})\approx 1$ (modest perturbation) preserving stability; when $\mathbf{x}$ deviates (e.g., OOD), $u(\mathbf{x})>1$ increases perturbation strength, amplifying sensitivity to atypical behavior \cite{wu2020adversarial, li2024revisiting}.

Practically, stronger perturbations on OOD inputs reveal regions whose gradient responses become erratic; we therefore (i) avoid throwing away informative, semantically consistent regions by not over-amplifying shallow noise, and (ii) penalize regions that exhibit excessive gradient fluctuation across stochastic realizations, which are likely spurious. Hyperparameters ($\alpha,\beta,\gamma,\rho_0$) offer direct control over global scale, input sensitivity and thresholding, while $\lambda$ in the covariance regularizer below stabilizes the Mahalanobis computation.

Over $T$ stochastic forward passes we collect activations $\{\mathbf{a}_k^{(t)}\}_{k=1..K}^{t=1..T}$ and compute layer-wise sensitivities
\begin{equation}
\eta_{\ell_k}^{(t)} \;=\; \big\| \nabla_{\mathbf{a}_k} \hat{y}^{(t)} \big\|_2.
\end{equation}
A descriptor per sample aggregates these sensitivities:
\begin{equation}
\mathbf{d}_i \;=\; \frac{1}{T}\sum_{t=1}^T \big[\,\eta_{\ell_1}^{(t)},\ldots,\eta_{\ell_K}^{(t)}\big].
\end{equation}
We quantify atypicality with a regularized Mahalanobis distance to the training descriptors $D_{\text{train}}$,
\begin{equation}
s_i \;=\; \sqrt{(\mathbf{d}_i - \mu)^\top (\Sigma + \lambda I)^{-1} (\mathbf{d}_i - \mu)},
\end{equation}
where $\mu=\mathrm{mean}(D_{\text{train}})$ and $\Sigma=\mathrm{cov}(D_{\text{train}})$. Normalizing $s_i$ across the batch yields
\begin{equation}
u_i \;=\; \frac{s_i - \min_j s_j}{\max_j s_j - \min_j s_j},
\qquad
s_{\text{conf}}(S) \;=\; 1 - u_i.
\end{equation}

\begin{algorithm}[t]
\caption{Greedy region selection for interpretable subset discovery}
\label{alg:greedy_alt}
\DontPrintSemicolon
\SetAlgoLined

\KwIn{Image $I \in \mathbb{R}^{w \times h \times 3}$, grid size $N \times N$, saliency $A$, divisions $m$, max subset size $k$}
\KwOut{Selected subset $S \subseteq V$, $|S| \leq k$}

$V \gets \emptyset$\;
$A \gets \text{resize}(A, N, N)$\;
$d \gets \frac{N \times N}{m}$\;

\For{$l \gets 1$ \KwTo $m$}{
    $I^{(l)} \gets I$\;
    \For{$i,j \in [1,N]$}{
        $r \gets \text{rank}(A, i, j)$\;
        \If{$r \le d(l-1)$ \textbf{or} $r > dl$}{
            Mask patch $(i,j)$ in $I^{(l)}$\;
        }
    }
    $V \gets V \cup \{I^{(l)}\}$\;
}

$S \gets \emptyset$\;
\For{$t \gets 1$ \KwTo $k$}{
    $\alpha \gets \arg\max_{u \in V \setminus S} F(S \cup \{u\})$\;
    $S \gets S \cup \{\alpha\}$\;
}

\Return{$S$}\;

\end{algorithm}

\subsection{Greedy Optimization and Formal Properties}

To select informative subsets of regions, we defined our objectives 
$F_{\text{attr}}(S)$ and $F_{\text{obj}}(S)$ as set functions over the ground set $V$. 
We adopt a greedy maximization scheme, where at each step the element 
with the highest marginal contribution is added.




Both objectives exhibit monotonicity and submodularity, 
making greedy selection well-suited for optimization.


\textbf{Theorem 1.} (\cite{nemhauser1978analysis})  
Let $S$ be the solution from the greedy algorithm and $S^*$ be the optimal solution. If $F(\cdot)$ is a non-negative, monotonic, and submodular function, the greedy solution is guaranteed to be near-optimal:
\begin{equation}
F(S) \geq \left( 1 - \frac{1}{e} \right) F(S^*)
\end{equation}
with $e$ the base of the natural logarithm.  
This establishes a constant-factor approximation bound for our greedy algorithm and ensures that our method
efficiently finds a high-quality, interpretable subset of regions
for both attribution and object-level explanation tasks.

\begin{table*}[t]
\centering
\scriptsize 
\rowcolors{2}{gray!10}{white} 
\begin{tabular}{llcccccccc}
\toprule
\textbf{Partition} & \textbf{Method} & 
\multicolumn{2}{c}{\textbf{CUB-200-2011 (ID)}} & 
\multicolumn{2}{c}{\textbf{NABirds (OOD - Related)}} & 
\multicolumn{2}{c}{\textbf{CIFAR-100 (OOD - Compl.)}} & 
\multicolumn{2}{c}{\textbf{CUB Transformed (OOD - Trans.)}} \\
\cmidrule(lr){3-4} \cmidrule(lr){5-6} \cmidrule(lr){7-8} \cmidrule(lr){9-10}
& & Ins. (↑) & Del. (↓) & Ins. (↑) & Del. (↓) & Ins. (↑) & Del. (↓) & Ins. (↑) & Del. (↓) \\
\midrule
Grad   & HSIC \cite{novello2022hsic}          & 0.6843 & 0.0647 & 0.3723 & 0.0342 & 0.1045 & 0.0447 & 0.1423 & 0.0245 \\
Grad   & HSIC + SMDL \cite{chen2024less}   & 0.6832 & 0.0321 & 0.4103 & 0.0239 & 0.1299 & 0.0442 & 0.1612 & 0.0191 \\
SEEDS  & HSIC + SMDL \cite{chen2024less}    & 0.6656 & 0.0241 & 0.4495 & 0.0159 & 0.3223 & 0.0297 & 0.3203 & 0.0167 \\
SEEDS  & HSIC + Ours     & \textbf{0.6991} & \textbf{0.0290} & \textbf{0.5110} & \textbf{0.0153} & \textbf{0.3549} & \textbf{0.0286} & \textbf{0.3312} & \textbf{0.0166} \\
SLICO  & HSIC + SMDL \cite{chen2024less}     & 0.7251 & 0.0272 & 0.5232 & 0.0170 & 0.3109 & 0.0256 & 0.3682 & 0.0176 \\
SLICO  & HSIC + Ours     & \textbf{0.7374} & \textbf{0.0266} & \textbf{0.5558} & \textbf{0.0163} & \textbf{0.3493} & \textbf{0.0256} & \textbf{0.3801} & \textbf{0.0173} \\
\hline
\end{tabular}
\caption{Comparison of Insertion AUC(↑) and Deletion AUC(↓) scores for different partition strategies on CUB-200-2011 (ID), NABirds (Related OOD), CIFAR-100 (Complementary OOD), and CUB-Transformed (Transformed OOD). Methods compared: HSIC-
Attribution vs. HSIC-
Attribution + SMDL vs. HSIC-Attribution + Ours}
\label{tab:Results for SLICO, SEEDS and Grad for CUB ID-OOD}
\end{table*}

\section{Experiments}

\subsection{Experimental Setup}

\paragraph{Datasets}
We designed two experimental settings to validate our hypothesis that subset selection–based methods for visual explainability exhibit degraded performance under out-of-distribution (OOD) conditions. Each setting consisted of an in-distribution (ID) dataset and three corresponding OOD datasets, following the taxonomy introduced in \cite{farquhar2022ood}. This taxonomy distinguishes meaningful categories of distribution shifts beyond the overly broad standard definition (i.e., any inequality between reference and target distributions). Specifically, our OOD datasets encompassed: (1) transformed-distribution, which involve applying well-defined functional alterations to the reference data, such as sensor noise, rotations, or adversarial perturbations (2) related-distribution, which capture contextually similar but shifted data based on deployment scenarios or task variations and (3) complementary distribution, which approximate the complement of the reference distribution by including unseen classes or input regions. These selections allowed us to systematically assess performance degradation across diverse shift types.

In the first setting, we adopted the CUB dataset, a fine-grained bird species classification dataset highlighted in \cite{chen2024less}, as the ID distribution. To construct OOD counterparts, we curated three variations. The related-distribution OOD was represented by the North American Birds (NABirds) dataset, which contains 555 species and exhibits strong distributional similarity to CUB. For the complementary-distribution OOD, we filtered the CIFAR-100 dataset to retain only 50 non-animal categories, ensuring minimal semantic overlap with CUB. Finally, to simulate transformed-distribution OOD, we perturbed the CUB dataset using additive noise and geometric transformations.

In the second setting, we considered COCO as the ID dataset. Here, the complementary-distribution OOD was provided by iNaturalist, a large-scale fine-grained species dataset with a taxonomic focus distinct from COCO. The related-distribution OOD was represented by the full CIFAR-100 dataset, which contains visually similar categories but differs in purpose and collection protocols. To generate the transformed-distribution OOD, we applied random blurs, rotations, and other perturbations to COCO images. 

\paragraph{Evaluation Metrics and Partition Strategies}
To assess the reliability of our approach, we focus on faithfulness, which captures how well the explanations align with the model’s decision process. We employ insertion and deletion AUC scores \cite{petsiuk2018rise} as evaluation metrics, measuring the change in predicted class probability as salient regions are progressively added or removed. Leveraging the greedy search strategy \ref{alg:greedy_alt}, region importance is naturally reflected by the sequence in which regions are explored. Results are reported for both ID and OOD settings to evaluate robustness in varied conditions. A key design element in our framework is the image partitioning strategy used to generate candidate subsets. We considered three partioning approaches. The first, grid-based partitioning guided by saliency \cite{simonyan2014deep}, divides images into fixed grids while emphasizing regions identified as salient, thus enabling structured selection of informative regions. The second, SLICO superpixels \cite{achanta2012slic}, adaptively groups pixels into perceptually coherent regions, offering a segmentation that aligns well with local image structure. The third, SEEDS superpixels \cite{van2012seeds}, refines partitions iteratively, allowing the resulting segments to adhere more closely to object boundaries. These strategies provided a spectrum of partitioning granularities and perceptual alignments, enabling a systematic study of their effect on subset quality across both ID and OOD conditions.
\vspace{-0.6cm}
\paragraph{Baseline \& Implementation Details}
In the first experimental setting, we selected HSIC-Attribution \cite{novello2022hsic} as the baseline method for obtaining prior saliency maps. We considered both the original formulation and its extension with Submodular Subset Selection (HSIC-Attribution + SMDL) \cite{chen2024less}. To assess the impact of our contributions, we further introduced a variant combining HSIC-Attribution with our proposed approach (HSIC-Attribution + Ours). In the second experimental setting, we evaluated our method in the context of object detection using the object-level foundational model GroundingDINO \cite{liu2024groundingdino}, and compared its performance against the existing Visual Precision Search(VPS) method \cite{chen2025visualprecision}. 

\section{Results}

\begin{table*}[ht]
\centering
\scriptsize 
\rowcolors{2}{gray!10}{white} 
\begin{tabular}{llcccccccc}
\toprule
\textbf{Partition} & \textbf{Method} & 
\multicolumn{2}{c}{\textbf{COCO-2017 (ID)}} & 
\multicolumn{2}{c}{\textbf{CIFAR-100 (OOD - Related)}} & 
\multicolumn{2}{c}{\textbf{iNaturalist (OOD - Compl.)}} & 
\multicolumn{2}{c}{\textbf{COCO Transformed (OOD - Trans.)}} \\
\cmidrule(lr){3-4} \cmidrule(lr){5-6} \cmidrule(lr){7-8} \cmidrule(lr){9-10}
& \textbf{GroundingDINO \cite{liu2024groundingdino}} & Ins. (↑) & Del. (↓) & Ins. (↑) & Del. (↓) & Ins. (↑) & Del. (↓) & Ins. (↑) & Del. (↓) \\
\midrule

SEEDS & VPS \cite{chen2025visualprecision} & 0.2878 & 0.0593 & 0.0688 & 0.0319 & 0.0880 & 0.0035 & 0.0673 & 0.0288 \\
SEEDS & Ours & \textbf{0.3056} & \textbf{0.0939} & \textbf{0.1308} & \textbf{0.0940} & \textbf{0.0982} & \textbf{0.0614} & \textbf{0.1406} & \textbf{0.0357} \\
SLICO & VPS \cite{chen2025visualprecision} & 0.3512 & 0.0725 & 0.0857 & 0.0276 & 0.0954 & 0.0349 & 0.0439 & 0.0300 \\
SLICO & Ours & \textbf{0.3729} & \textbf{0.1256} & \textbf{0.1239} & \textbf{0.1054} & \textbf{0.1028} & \textbf{0.0582} & \textbf{0.0879} & \textbf{0.0258} \\
\bottomrule
\end{tabular}
\caption{Comparison of Insertion AUC(↑) and Deletion AUC(↓) scores for different partition strategies on COCO (In-Distribution), COCO-Transformed (Transformed OOD), CIFAR-100 (Related OOD), and iNaturalist (Complementary OOD). Methods compared: Visual Precision Search vs. Ours on object-level foundational model GroundingDINO for object detection tasks}
\label{tab:Results for SLICO and SEEDS for COCO ID-OOD}
\end{table*}

\begin{figure} 

\begin{minipage}{0.495\textwidth}
    \includegraphics[width=\linewidth]{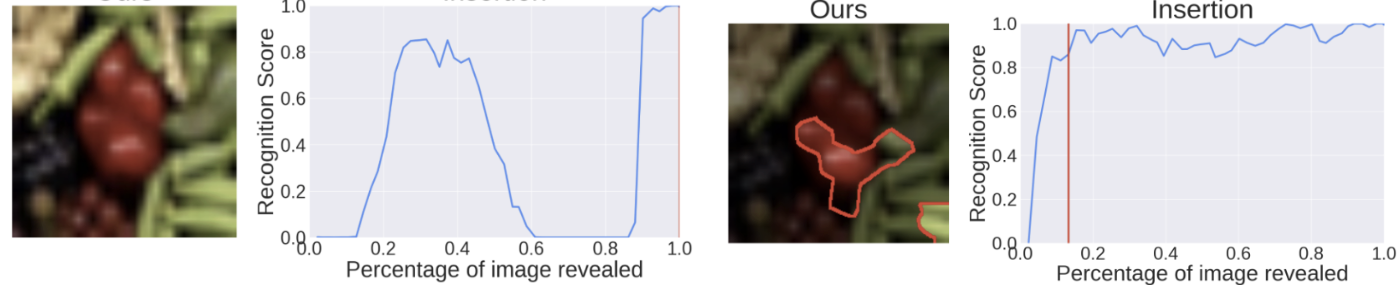}
\end{minipage}

\begin{minipage}{0.495\textwidth}
    \includegraphics[width=\linewidth]{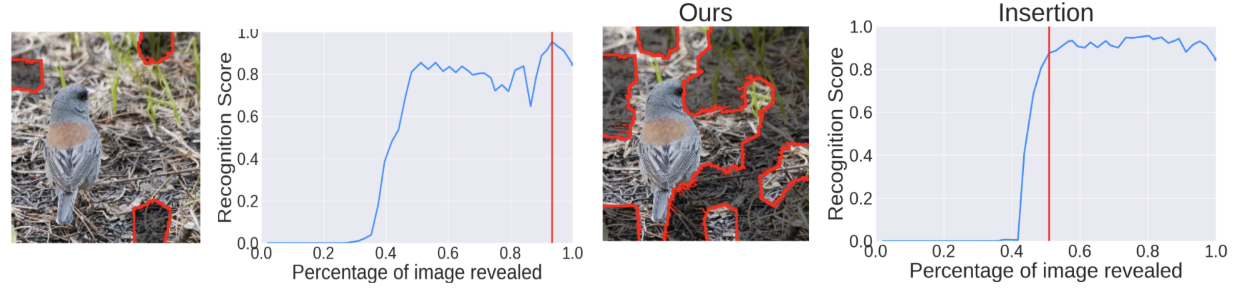}
\end{minipage}

\caption{\textbf{Qualitative comparison on OOD samples}: The figure contrasts the baseline HSIC+SMDL (left) with our proposed method (right) on a 'sweet pepper' from CIFAR-100 (top) and a bird from North American Birds (bottom).}
\label{fig:set1}
\end{figure}

\begin{figure} 

\begin{subfigure}{0.495\textwidth}
    \includegraphics[width=\linewidth]{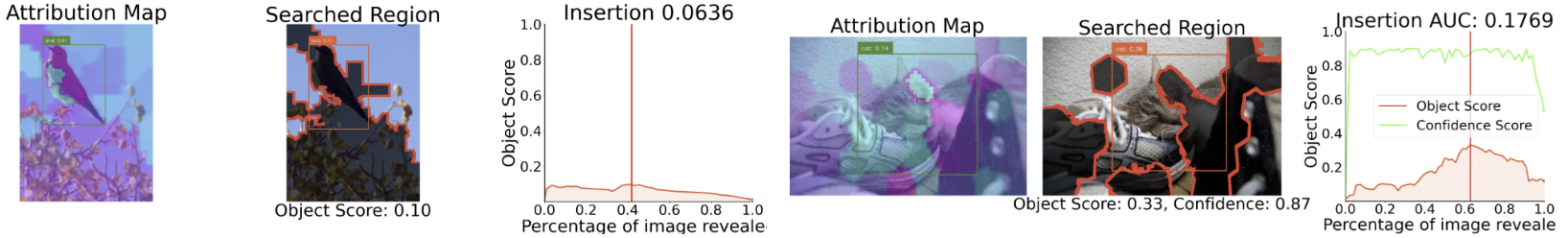}
\end{subfigure}

\begin{subfigure}{0.495\textwidth}
    \includegraphics[width=\linewidth]{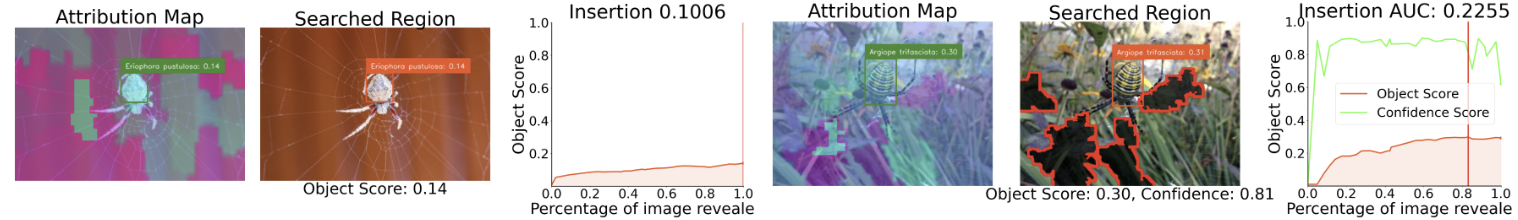}
\end{subfigure}

\caption{\textbf{Qualitative comparison for object-level interpretation on OOD examples}: This figure contrasts the baseline VPS method (left) with our proposed method (right) on a transformed COCO image (top) and an iNaturalist image (bottom)}
\label{fig:set2}
\end{figure}

\subsection{Performance Summary}




Table~\ref{tab:Results for SLICO, SEEDS and Grad for CUB ID-OOD} reports Insertion AUC (higher is better) and Deletion AUC (lower is better) across in-distribution (CUB-200-2011) and multiple OOD settings for the fine-grained classification task. Our method consistently improves Insertion while reducing or maintaining Deletion, indicating stronger attribution fidelity and stability. On CUB-200-2011, SLICO improves from 0.7251 to 0.7374 (+1.7\%) with a corresponding Deletion drop from 0.0272 to 0.0266 (–2.2\%), while SEEDS achieves a +5.0\% Insertion gain (0.6656 → 0.6991).

The improvements are more pronounced under OOD shifts. On NABirds, Insertion rises by +6.2\% (SLICO) and +13.7\% (SEEDS), with consistent Deletion reductions of 3–5\%. On CIFAR-100, which poses a complementary OOD challenge, Insertion increases by +12.3\% (SLICO) and +10.1\% (SEEDS), while Deletion also decreases (e.g., 0.0297 → 0.0286 for SEEDS). Even in transformed OOD settings, such as CUB Transformed, gains are steady: +3.2\% (SLICO) and +3.4\% (SEEDS) in Insertion with marginally lower Deletion.

Table~\ref{tab:Results for SLICO and SEEDS for COCO ID-OOD}  details the performance of our method against the Visual Precision Search (VPS) baseline on object detection tasks using the GroundingDINO model. The results show that our approach yields substantial improvements in Insertion AUC across all conditions, indicating a significantly enhanced ability to identify critical object features. However, this often comes with a trade-off, as Deletion AUC scores are generally higher.

On the in-distribution COCO dataset, our method boosts Insertion AUC by +6.2\% for both SLICO (0.3512 → 0.3729) and SEEDS (0.2878 → 0.3056) partitions. The most dramatic gains are observed under OOD shifts. For instance, on the related OOD dataset CIFAR-100, our method more than doubles the Insertion AUC, with a +44.5\% increase for SLICO (0.0857 → 0.1239) and a +80.0\% increase for SEEDS (0.0688 → 0.1308). Similarly, on the transformed COCO dataset, Insertion AUC improves by +100.2\% (SLICO) and +108.9\% (SEEDS). While the Deletion AUC is higher in most scenarios, our method achieves a -14.0\% reduction in Deletion AUC (0.0300 → 0.0258) with the SLICO partition on transformed OOD data, demonstrating its potential for stability under certain conditions.


To visually validate our quantitative results, Figures~\ref{fig:set1} and~\ref{fig:set2} provide qualitative comparisons on challenging OOD examples for fine-grained classification and object-level interpretation, respectively.

In the classification task shown in figure~\ref{fig:set1}, our uncertainty-aware framework identifies more compact and semantically coherent subsets than the baseline, focusing directly on the object of interest rather than fragmented or background regions. This improved precision translates to more efficient and stable explanations, as evidenced by the steeper rise in the Insertion score curves for our method.

This enhanced robustness extends to object-level interpretation~\ref{fig:set2}, where our method successfully identifies a collection of patches that more accurately delineates the target object. This leads to a marked improvement in both Object Scores and Insertion AUCs compared to the baseline approach.

Across both distinct tasks, this visual evidence reinforces a consistent finding. The precisely highlighted regions selected by our method serve as a direct visual confirmation of its superior attribution quality, showing a clear focus on semantically relevant features. By integrating an uncertainty-aware confidence score, our framework produces significantly more precise, robust, and interpretable explanations, particularly when faced with distributional shifts.

\subsection{Discovering the prediction on OOD scenarios}

Gradient-based uncertainty estimation adapts differently across OOD types, while consistently following the perturbation–gradient–Mahalanobis pipeline in Algorithm~\ref{alg:uncertainty}  

For \textbf{related OOD} (e.g., CUB $\rightarrow$ NABirds), inputs remain semantically close but shift in feature statistics. The modulation term 
$u(\mathbf{x})$ rises moderately, amplifying subtle layer-specific deviations. Perturbations accentuate these shifts, gradient norms $\eta_{\ell_k}^{(t)}$ identify weakly aligned layers, and the aggregated descriptor $\mathbf{d}_i$ departs selectively from the training cloud. The Mahalanobis distance emphasizes these low-variance directions, enabling reliable detection of domain-shifted yet related inputs.  

For \textbf{complementary OOD} (e.g., CUB $\rightarrow$ CIFAR-100 non-animals), features $\phi_\ell(\mathbf{x})$ lie far from the centroid $\bar{\phi}$, driving $u(\mathbf{x})$ to high values. Perturbed weights $\tilde{\theta}_\ell^{(t)}$ produce highly variable outputs, and gradient norms remain large across layers. The resulting descriptor $\mathbf{d}_i$ is displaced significantly, yielding a large Mahalanobis distance $s_i$. Rather than harming interpretability, this high uncertainty penalizes unstable regions in the submodular objective, guiding subset selection toward consistent, semantically informative evidence.  

For \textbf{transformed OOD} (e.g., blur, noise, distortions of ID data), deviations in $\phi_\ell(\mathbf{x})$ are moderate and concentrated in early, low-level layers. $u(\mathbf{x})$ scales proportionally, amplifying gradients primarily in those layers. Aggregated descriptors $\mathbf{d}_i$ therefore shift systematically from ID descriptors, and the Mahalanobis distance grows in affected dimensions while remaining bounded elsewhere. This yields calibrated uncertainty that distinguishes corrupted ID samples from clean ones.  

\section{Conclusion}

In this work we introduced a lightweight, uncertainty-aware submodular algorithm that closes a critical robustness gap in visual explanations under distribution shifts. Our results demonstrate that integrating principled uncertainty via adaptive gradient perturbations yields more compact and stable explanations for both in-distribution (ID) and out-of-distribution (OOD) data.

While our approach shows a benefit, exploring the trade-off between attribution fidelity and the computational cost of stochastic estimation presents an interesting optimization challenge. Refining the nuanced balance between Insertion and Deletion metrics in object detection is another promising direction.

\end{document}

%% file: preamble.tex
\usepackage[T1]{fontenc}
\usepackage{lmodern}  
\usepackage{lipsum} 

\usepackage{amsmath}
\usepackage{amssymb}
\usepackage{amsfonts}
\usepackage{amsthm}

\usepackage[linesnumbered,ruled,vlined]{algorithm2e}

\usepackage{graphicx}
\usepackage{epsfig}
\usepackage{caption}
\usepackage{subcaption}
\usepackage[table]{xcolor}
\usepackage{booktabs} 
\usepackage{subcaption}

\usepackage{times}
\usepackage{microtype} 
\usepackage{xcolor}    
\usepackage{enumitem}  
\usepackage{nicefrac}  
\usepackage{siunitx}   

\usepackage[numbers]{natbib} 
\usepackage{url}

\usepackage[pagebackref,breaklinks,colorlinks]{hyperref}
\hypersetup{
    colorlinks=true,
    linkcolor=blue,
    citecolor=blue,
    filecolor=magenta,
    urlcolor=blue,
    pdfauthor={Your Name},
    pdftitle={Your Paper Title},
    pdfsubject={Research Paper},
    pdfkeywords={keywords, here}
}
\usepackage[accsupp]{axessibility}  

